\documentclass[letterpaper, 10 pt, journal, twoside]{IEEEtran}
\ifCLASSINFOpdf
\else
\fi
\usepackage{graphicx} 
\usepackage{epsfig} 
\usepackage{dblfloatfix} 
\usepackage{courier}
\usepackage{amsmath} 
\usepackage{amssymb}
\usepackage{float}
\usepackage[backend=bibtex,bibstyle=ieee,citestyle=numeric-comp]{biblatex}
\usepackage{balance}

\newcommand{\ignore}[1]{}

\newcommand{\norm}[1]{\left\Vert#1\right\Vert} 
\newcommand{\mc}[1]{\mathcal{#1}}

\newcommand{\bma}[1]{\left[\begin{array}{ #1}}
\newcommand{\ema}{\end{array}\right]}

\DeclareMathAlphabet{\mbf}{OT1}{ptm}{b}{n}
\newcommand{\mbs}[1]{{\boldsymbol{#1}}}

\newcommand{\mbfbar}[1]{{\bar{\mbf{#1}}}}

\def\fdotb{{\raisebox{-0.6ex}{ \kern0.2ex\raisebox{0.8ex}{\tiny $\hspace*{-1ex}\circ$}}}}
\def\fddotb{{\raisebox{-0.6ex}{ \kern0.2ex\raisebox{0.8ex}{\tiny $\hspace*{-1ex}\circ\circ$}}}}

\newcommand{\p}{\partial}
\newcommand{\f}{\frac}

\newcommand{\trans}{{\ensuremath{\mathsf{T}}}} 
\newcommand{\utimes}{ {\raisebox{-0.6ex}{ \kern-1.0ex\raisebox{0.6ex}{ \small $\mathsf{v}$}}} } %
\newcommand{\imag}{ {\ensuremath{\mathrm{Im}}} } 

\newcommand{\beq}{\begin{equation}}
\newcommand{\eeq}{\end{equation}}
\newcommand{\bdis}{\begin{displaymath}}
\newcommand{\edis}{\end{displaymath}}
\newcommand{\beqarray}{\begin{eqnarray}}
\newcommand{\eeqarray}{\end{eqnarray}}
\newcommand{\beqarraynn}{\begin{eqnarray*}}
\newcommand{\eeqarraynn}{\end{eqnarray*}}
\newcommand{\balign}{\begin{align}}
\newcommand{\ealign}{\end{align}}
\newcommand{\balignnn}{\begin{align*}}
\newcommand{\ealignnn}{\end{align}}

\renewcommand{\theenumii}{\arabic{enumii}}
\makeatletter
\renewcommand{\p@enumii}{\theenumi.}
\makeatother

\long\def\isdef{\mathrel{\mathop{\smash{=}}\limits^{\hbox{\raisebox{.7ex}{\small$\scriptstyle\triangle
$}}}}}

\begin{document}
%
%
%
%
%
%
%
\def \myJournal {IEEE Robotics and Automation Letters}
\def \myDoi {10.1109/LRA.2020.2965882}
\def \myPaperSiteName {IEEE Xplore}
\def \myPaperSiteLink {https://ieeexplore.ieee.org/document/8957301}
\def \myYear {2020}
\def \myPaperCitation{C. C. Cossette, A. Walsh and J. R. Forbes, ``The Complex-Step Derivative Approximation on Matrix Lie Groups," in \textit{IEEE Robotics and Automation Letters}, vol. 5, no. 2, pp. 906-913, April 2020.}


\begin{figure*}[t]

\thispagestyle{empty}
\begin{center}
\begin{minipage}{6in}
\centering
This paper has been accepted for publication in \emph{\myJournal}. 
\vspace{1em}

This is the author's version of an article that has, or will be, published in this journal or conference. Changes were, or will be, made to this version by the publisher prior to publication.
\vspace{2em}

\begin{tabular}{rl}
DOI: & \myDoi\\
\myPaperSiteName: & \texttt{\myPaperSiteLink}
\end{tabular}

\vspace{2em}
Please cite this paper as:

\myPaperCitation

\vspace{15cm}
\copyright \myYear \hspace{4pt}IEEE. Personal use of this material is permitted. Permission from IEEE must be obtained for all other uses, in any current or future media, including reprinting/republishing this material for advertising or promotional purposes, creating new collective works, for resale or redistribution to servers or lists, or reuse of any copyrighted component of this work in other works.

\end{minipage}
\end{center}
\end{figure*}
\newpage
\clearpage
\pagenumbering{arabic} 
\title{The Complex-Step Derivative Approximation on Matrix Lie Groups}
%
%
%

\author{Charles Champagne Cossette$^{1}$, Alex Walsh$^{2}$, and James Richard Forbes$^{3}$
\thanks{Manuscript received: Sept. 10, 2019; Revised Nov. 28, 2019; Accepted Dec. 23, 2019. This paper was recommended for publication by Editor Sven Behnke upon evaluation of the Associate Editor and Reviewers' comments. This work was supported by NSERC Collaborative Research and Development, Engage, and Discovery Grant programs} 
\thanks{$^{1}$Ph.D. Candidate, Department of Mechanical Engineering, McGill University, 817 Sherbrooke St. W., Montreal, QC,
Canada, H3A 0C3. e-mail: {\tt\small charles.cossette@mail.mcgill.ca}}%
\thanks{$^{2}$Postdoctoral Researcher, Department of Mechanical Engineering, McGill University, 817 Sherbrooke St. W., Montreal, QC,
Canada, H3A 0C3. e-mail: 
        {\tt\small alex.walsh@mail.mcgill.ca}}%
\thanks{$^{3}$Associate Professor, Department of Mechanical Engineering, McGill University, 817 Sherbrooke St. W., Montreal, QC,
Canada, H3A 0C3. e-mail: 
        {\tt\small james.richard.forbes@mcgill.ca}}
\thanks{Digital Object Identifier (DOI): see top of this page.}
\vspace{-1cm}}

%
%

\markboth{IEEE Robotics and Automation Letters. Preprint Version. Accepted December, 2019}
{Cossette \MakeLowercase{\textit{et al.}}: The Complex-Step Derivative Approximation on Matrix Lie Groups} 

%



\maketitle

\begin{abstract}
The complex-step derivative approximation is a numerical differentiation technique that can achieve analytical accuracy, to machine precision, with a single function evaluation. In this paper, the complex-step derivative approximation is extended to be compatible with elements of matrix Lie groups. As with the standard complex-step derivative, the method is still able to achieve analytical accuracy, up to machine precision, with a single function evaluation. Compared to a central-difference scheme, the proposed complex-step approach is shown to have superior accuracy. The approach is applied to two different pose estimation problems, and is able to recover the same results as an analytical method when available.
\end{abstract}


%
\IEEEpeerreviewmaketitle

\section{Introduction}
%
%
%
%
\IEEEPARstart{A}{ttitude} and pose, ubiquitous entities of interest in robotics problems, are most naturally represented as elements of matrix Lie groups. Path planning, state estimation, and control algorithms often require Jacobian computations with respect to attitude and pose. Often these Jacobians are computed analytically, by hand, via a Taylor-series expansion while adhering to the matrix Lie group structure of the problem \cite{Barfoot2019}. However, in some cases analytical computation of Jacobians may be impractical, necessitating a numerical procedure. 
%
%
Numerical computation of Jacobians is also useful for quickly comparing algorithms that require Jacobians, before investing effort into one specific algorithm and the associated analytically derived Jacobians.
Numerical Jacobians can also be used to verify Jacobians that are derived by hand.
%

A variety of numerical differentiation techniques appropriate for matrix Lie groups can be found in the literature.  In \cite{Absil2008} a forward-difference method is described for general matrix manifolds, a method that is used in the open-source software MANOPT \cite{Boumal2014}. A central-difference method is employed in the open-source software GTSAM \cite{GTSAM2019a}, and algorithmic differentiation methods are presented in \cite{Robenack2011,Sommer2013}. The Python-based software PYMANOPT \cite{Townsend2016} is an open-source optimization toolbox for matrix manifolds that employs algorithmic differentiation. SOPHUS \cite{Sophus2019} is another open-source C++ package that exploits the automatic differentiation functionality available in CERES \cite{Ceres2019}, a nonlinear least-squares library developed by Google. However, algorithmic differentiation can be time consuming to implement and finite-differencing is prone to subtractive cancellation errors, thus limiting precision \cite{Martins2003}. The complex-step derivative approximation is a numerical method for computing first derivatives that does not suffer from subtractive cancellation errors \cite{Martins2003}. One of the earlier appearances of the complex-step derivative can be found in \cite{Squire1998}, where the derivatives of scalar functions of real variables are evaluated. In \cite{Martins2003}, the complex-step derivative is investigated further, along with its use in Fortran, C/C++, and other languages. An application to a multidisciplinary design optimization problem is also shown. This method has gained popularity due to its ability to realize machine-precision accuracy of derivative computations, and doing so without tuning the step size, since it can be reduced to an arbitrarily small value. The complex-step derivative also requires only one complex function evaluation, which is beneficial compared to central-differencing when the function is expensive to evaluate. The complex-step derivative is straight-forward to implement, especially in \textsc{Matlab}, where the default variable type is complex.

This paper considers the formulation and application of the complex-step derivative approximation to functions of matrix Lie group elements.  The aforementioned advantages of the standard complex-step derivative remain present, while the proposed method can be used to compute both left and right Jacobians. Various examples are presented, demonstrating the utility and advantages of the matrix Lie group version of the complex-step derivative. In particular, pose estimation problems are considered, one using the ETH Z\"urich EuRoC dataset \cite{Burri2016}, where analytical Jacobians are available for comparison, and one using the `Lost in the Woods' dataset \cite{Barfoot2009}, where computation of analytical Jacobians is possible, but time consuming. When solving for the maximum a posteriori (MAP) estimate of the pose using a Gauss-Newton algorithm, it is shown that computing the Jacobians using the complex-step derivative realizes the same accuracy and convergence properties as when analytical Jacobians are used.

\vspace{-5pt}
\section{Preliminaries}

\subsection{Matrix Lie Groups}
A matrix Lie group $\mc{G}$ is a Lie group that consists of the set of $m \times m$ invertible matrices, where the group operation is matrix multiplication \cite[Ch.~10.2]{Chirikjian2009}. From the definition of a group, a matrix Lie group is closed under matrix multiplication. That is, given $\mbf{X},\mbf{Y} \in \mc{G}$, it follows that $\mbf{XY} \in \mc{G}$. A matrix Lie group is a closed subgroup of the general linear group defined by \cite[Ch.~1.1]{Hall2015}
\bdis
GL(m,\mathbb{C}) = \{ \mbf{X} \in \mathbb{C}^{m\times m} \;|\; \det (\mbf{X}) \neq 0 \},
\edis
which is also a matrix Lie group. The matrix {Lie algebra} of $\mc{G}$ is denoted $\mathfrak{g}$, and is defined as \cite[Ch.~3.3]{Hall2015},
\beq \label{eq:lieAlgebra}
\mathfrak{g} = \{ \mbs{\Xi} \;|\; \exp(t\mbs{\Xi}) \in \mc{G}, \forall t \in \mathbb{R} \}.
\eeq
It can be shown that the matrix Lie algebra defined by \eqref{eq:lieAlgebra} is a valid Lie algebra \cite[Ch.~3.1]{Hall2015}, and is a vector space closed under the operation of the {Lie bracket} $[\cdot,\cdot]$, which can be computed by
$
[\mbf{A}, \mbf{B}] = \mbf{A}\mbf{B} - \mbf{B} \mbf{A} \in \mathfrak{g},$ for all  $\mbf{A}, \mbf{B} \in \mathfrak{g}.
$
The wedge operator $(\cdot)^\wedge: \mathbb{R}^n \to \mathfrak{g}$ maps a column matrix to the matrix Lie algebra. The exponential map $\exp(\cdot):\mathfrak{g} \to \mc{G}$ maps an element of the matrix Lie algebra to the matrix Lie group, and is computed using the matrix exponential. The only matrix Lie group elements $\mbf{X} \in \mc{G}$ that are considered in this paper are those that can be written as
$$
\mbf{X} = \exp(\mbs{\xi}^\wedge),
$$
where ${\mbs{\xi} \in \mathbb{R}^n}$. The ``vee'' operator ${(\cdot)^\vee : \mathfrak{g} \to \mathbb{R}^n}$ maps an element of the matrix Lie algebra to a column matrix. The logarithmic map $\ln(\cdot):\mc{G} \to \mathfrak{g}$ maps an element of the matrix Lie group to the matrix Lie algebra, and is computed by the matrix logarithm. A parameterization of the group $\mc{G}$ can be retrieved from $\mbf{X}$ via
$$
\mbs{\xi} = \ln(\mbf{X})^\vee,
$$
when the matrix logarithm is well defined. The adjoint representation of $\mbf{X}$ is denoted $\mathrm{Ad}(\mbf{X})$, such that ${(\mathrm{Ad}(\mbf{X})\mbs{\zeta})^\wedge = \mbf{X} \mbs{\zeta}^\wedge \mbf{X}^{-1}},\; \mbs{\zeta} \in \mathbb{R}^n$. This leads to the identity
\bdis
\exp((\mathrm{Ad}(\mbf{X}) \mbs{\zeta})^\wedge) = \mbf{X} \exp(\mbs{\zeta}^\wedge) \mbf{X}^{-1}.
\edis
The {Baker-Campbell-Hausdorff} (BCH) formula is the solution to
\bdis
\mbf{z} = \ln (\exp(\mbs{\xi}_1^\wedge) \exp (\mbs{\xi}_2^\wedge)),
\edis
and the exact solution is an infinite sum \cite[Ch.~7.1.5]{Barfoot2019}. A first-order approximation to the BCH formula is
\bdis
  \ln (\exp(\mbs{\xi}_1^\wedge) \exp (\mbs{\xi}_2^\wedge)) = \mbs{\xi}_1^\wedge + \mbs{\xi}_2^\wedge,
\edis
which is exact in the event that $[\mbs{\xi}_1^\wedge, \mbs{\xi}_2^\wedge] = \mbf{0}$. Such an approximation is typically used when both $\mbs{\xi}_1$ and $\mbs{\xi}_2$ are assumed to be small. The details of the special Euclidean groups $SE(2)$, $SE(3)$, and the group of double direct isometries $SE_2(3)$ can be found in the appendix.


\subsection{Gauss-Newton Algorithm}
The {Gauss-Newton} algorithm is an optimization algorithm appropriate for nonlinear least-squares functions of the form
\beq \label{eq:gaussNewton}
J(\mbf{x}) = \frac{1}{2} \mbf{e}(\mbf{x})^\trans \mbf{W} \mbf{e} (\mbf{x}),
\eeq
where $\mbf{W} \in \mathbb{R}^{q \times q}$ is a symmetric positive definite weight matrix and ${\mbf{e}:\mathbb{R}^p \to \mathbb{R}^q}$ is some error function. Employing Newton's method directly on \eqref{eq:gaussNewton} requires the Hessian of $J(\mbf{x})$, which is potentially difficult to obtain. An alternate strategy is to substitute a first-order approximation of $\mbf{e}(\mbf{x})$ about some nominal $\mbfbar{x}$, given by \cite[Ch.~4.3]{Barfoot2019}
\bdis
\mbf{e}(\mbfbar{x} + \delta \mbf{x}) \approx \mbf{e}(\mbfbar{x}) + \left. \f{\p \mbf{e}(\mbf{x})}{\p \mbf{x}} \right|_{\mbf{x} = \mbfbar{x}} \delta \mbf{x},
\edis
into \eqref{eq:gaussNewton}, thus yielding the Jacobian and a Hessian approximation of $J(\mbf{x})$,
\begin{multline*}
J(\mbf{x}) \approx \frac{1}{2} \mbf{e}(\mbfbar{x})^\trans \mbf{W} \mbf{e} (\mbfbar{x})\\
 +  \underbrace{\mbf{e}(\mbfbar{x})^\trans\mbf{W}  \frac{\p \mbf{e}(\mbf{x})}{\p \mbf{x}}}_{\left.\frac{\p J(\mbf{x})}{\p \mbf{x}}\right|_{\mbfbar{x}}} \delta \mbf{x} + \frac{1}{2}\delta \mbf{x}^\trans \underbrace{\left(\frac{\p \mbf{e}(\mbf{x})}{\p \mbf{x}} \right)^\trans \mbf{W} \left(\frac{\p \mbf{e}(\mbf{x})}{\p \mbf{x}} \right)}_{\left.\frac{\p J(\mbf{x})}{\p \mbf{x} \p \mbf{x}^\trans}\right|_{\mbfbar{x}}} \delta \mbf{x}.
\end{multline*}
The Gauss-Newton algorithm then proceeds identically to Newton's method, where the nominal point is iterated by $\mbfbar{x}_{\ell} = \mbfbar{x}_{\ell-1} + \delta \mbf{x}_{\ell-1}$. The step $\delta\mbf{x}_{\ell-1}$ is calculated as
\bdis
\delta \mbf{x}_{\ell-1} = -\left(\left.\frac{\p J(\mbf{x})}{\p \mbf{x} \p \mbf{x}^\trans}\right|_{\mbfbar{x}_{\ell-1}}\right)^{-1}\left(\left.\frac{\p J(\mbf{x})}{\p \mbf{x}}\right|_{\mbfbar{x}_{\ell-1}}\right)^\trans.
\edis
\section{The Complex-Step Derivative Approximation}
\subsection{Review}
Consider the complex-differentiable function ${f:\mathbb{C}\to \mathbb{C}}$ perturbed about the nominal point $\bar{x}$ by $jh$ where $\bar{x},h \in \mathbb{R}$ and $j = \sqrt{-1}$. A Taylor series expansion yields
\begin{multline} \label{eq:complex1}
f(\bar{x} + jh) = f(\bar{x}) + \left. \frac{\p f(z)}{\p z} \right\rvert_{z = \bar{x}} jh \\ - \frac{1}{2} \left. \frac{\p^2 f(z)}{\p z^2} \right\rvert_{z = \bar{x}}h^2 - \frac{1}{3!} \left. \frac{\p^3 f(z)}{\p z^3} \right\rvert_{z = \bar{x}}jh^3  \ldots
\end{multline}
If $f(\bar{x})$ is assumed to be real for all real $\bar{x}$, then, to first order, taking the imaginary portion of \eqref{eq:complex1} yields \cite{Squire1998}
\bdis
 \left. \frac{\p f(z)}{\p z} \right\rvert_{z = \bar{x}} = \frac{\imag \{f(\bar{x} + jh)\}}{h} + \mc{O}(h^2).
 \edis
This is valid as long as $f(\bar{x}) \in \mathbb{R}$ for all $\bar{x} \in \mathbb{R}$, and that derivatives are evaluated at strictly real nominal points. From a practical standpoint, a user is often attempting to find derivatives of $f: \mathbb{R} \to \mathbb{R}$. Providing that this can be extended to $f:\mathbb{C}\to \mathbb{C}$ such that $f$ is complex-differentiable, then with a minor abuse of notation, this can construct a derivative approximation for $f(x)$ as written in {\cite{Squire1998,Martins2003}},

\bdis
\frac{\p f(x)}{\p x} \approx \frac{\imag \{f(x + jh)\}}{h}.
\edis
Since there are no subtractive cancellation errors, the complex-step derivative approximation can produce machine-precision approximations by reducing $h$ to an arbitrarily small step size.

\subsection{The Complex-Step Derivative on Matrix Lie Groups}
Consider a complex-differentiable function ${f:\mc{G} \to \mathbb{C}}$ where ${\mc{G} \subset GL(m,\mathbb{C})}$, ${\mbf{X}(\mbs{\epsilon}^R) = \mbfbar{X} \exp(\mbs{\epsilon}^{R^\wedge})}$ is parametrizable by a perturbation ${\mbs{\epsilon}^R = [ \epsilon_1^R\; \epsilon_2^R \ldots \epsilon_n^R]^\trans \in \mathbb{C}^n}$ on the right, and $\mbfbar{X} \in \mathbb{R}^{m \times m}$ is some nominal value of $\mbf{X}$. Consider perturbing $f(\mbf{X}(\mbs{\epsilon}^R))$ by $\mbs{\epsilon}^R = \mbf{0} + jh\mbf{1}_i$, where $\mbf{1}_i$ is the $i^\mathrm{th}$ column of the appropriately-dimensioned identity matrix $\mbf{1}$. The composition $f(\mbf{X}(\mbs{\epsilon}^R))$ has essentially recast $f$ as $f:\mathbb{C}^n \to \mathbb{C}$, from which a Taylor series expansion yields \cite[Ch.~11.3]{Chirikjian2009}
\begin{multline}\label{eq:complex2}
f\left(\mbfbar{X}\exp((jh\mbf{1}_i)^\wedge)\right) = f(\mbfbar{X}) + \left. \frac{\p f(\mbf{X}(\mbs{\epsilon}^R))}{\p \epsilon_i^R} \right\rvert_{\mbs{\epsilon}^R = \mbf{0}} jh \\ - \frac{1}{2} \left. \frac{\p^2 f(\mbf{X}(\mbs{\epsilon}^R))}{\p \epsilon_i^{R^2}} \right\rvert_{\mbs{\epsilon}^{R} = \mbf{0}}h^2 
 + \mc{O}(h^3).
\end{multline}
Since it is assumed that $f(\mbfbar{X}) \in \mathbb{R}$, taking the imaginary component of \eqref{eq:complex2} yields an approximation for the derivative
\beq \label{eq:complex_lie}
\frac{\p f(\mbf{X}(\mbs{\epsilon}^R))}{\p \epsilon_i^R} \approx \frac{\imag \{f\left(\mbfbar{X}\exp((jh\mbf{1}_i)^\wedge)\right)\}}{h}.
\eeq
The right Jacobian $\p f(\mbf{X}(\mbs{\epsilon}^R))/\p \mbs{\epsilon}^R$ can be obtained by individually computing the derivatives using \eqref{eq:complex_lie} with ${i = 1,2,\ldots ,n}$. The left Jacobian can identically be obtained by instead parametrizing $\mbf{X}$ with $\mbf{X}(\mbs{\epsilon}^{L}) = \exp(\mbs{\epsilon}^{L^\wedge})\mbfbar{X}$. This leads to
\beq \label{eq:complex_lie2}
\frac{\p f(\mbf{X}(\mbs{\epsilon}^L))}{\p \epsilon_i^L} \approx \frac{\imag \{f\left(\exp((jh\mbf{1}_i)^\wedge)\mbfbar{X}\right)\}}{h}.
\eeq 
Note that the superscripts on $\mbs{\epsilon}^R$ and $\mbs{\epsilon}^L$ are simply labels that correspond to right and left perturbations, respectively, as opposed to exponents.

\begin{figure}[t]
      \centering
      \makebox{\includegraphics[width = 1.05\columnwidth]{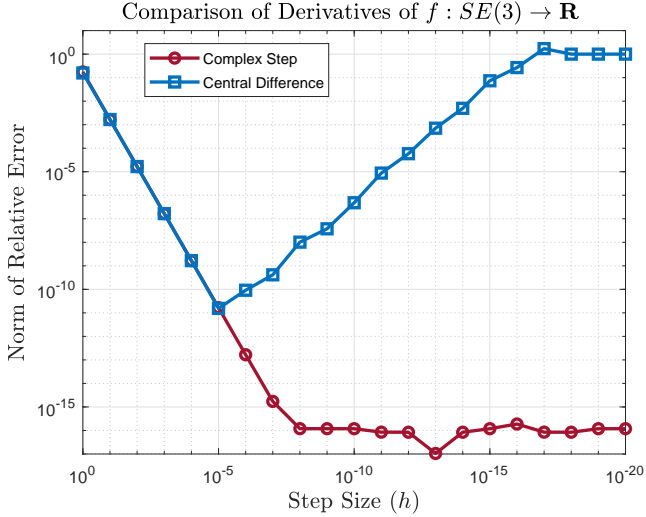}}
      \caption{Variation of relative error in gradient of $f:SE(3) \to \mathbb{R}$ with step size, for both complex-step and central-difference methods. Machine precision is achievable with a sufficient reduction in step size.\vspace{-0.5cm} }
      \label{fig:complexStepCompSE3}
\end{figure} 
\emph{Example 1:} Consider the function
\bdis
f(\mbf{T}) = \mbf{v}^\trans \mbf{T} \mbf{y},
\edis
where $\mbf{T} \in SE(3)$ and $\mbf{v}, \mbf{y} \in \mathbb{R}^4$. The left Jacobian can be determined analytically using the first-order approximation $\mbf{T} = \exp(\mbs{\epsilon}^{L^\wedge}) \mbfbar{T} \approx ( \mbf{1} + \mbs{\epsilon}^{L^\wedge})\mbfbar{T}$ and a Taylor series expansion. To this end,
\begin{align}
f(\exp(\mbs{\epsilon}^{L^\wedge})\mbfbar{T}) &=  \mbf{v}^\trans \exp(\mbs{\epsilon}^{L^\wedge})\mbfbar{T} \mbf{y} \nonumber \\
&\approx \mbf{v}^\trans ( \mbf{1} + \mbs{\epsilon}^{L^\wedge})\mbfbar{T} \mbf{y} \nonumber\\
&= \mbf{v}^\trans \mbfbar{T} \mbf{y} + \hspace{-8pt} \underbrace{\mbf{v}^\trans (\mbfbar{T} \mbf{y})^\odot}_{\left.\frac{\p f(\mbf{T}(\mbs{\epsilon}^L))}{\p \mbs{\epsilon}^L}\right|_{\mbs{\epsilon}^L = \mbf{0}}} \hspace{-8pt} \mbs{\epsilon}^L, \label{eq:jac_se3}
\end{align}
where the $(\cdot)^\odot$ operator is defined in the appendix. The elements of $\p f(\mbf{T}(\mbs{\epsilon}^L))/\p \mbs{\epsilon}^L$ are computed using \eqref{eq:complex_lie2} with varying step sizes $h$, and the results are compared with a central-difference scheme in Fig. \ref{fig:complexStepCompSE3}. The error is computed by taking the relative 2-norm of the difference between the analytical and numerical solutions. Like the standard complex-step derivative, the complex-step derivative  tailored to the matrix Lie group $SE(3)$ is able to achieve analytic accuracy, up to machine precision, for small enough $h$, while the central-difference derivative is not.  \\


\begin{figure}[t]
      \centering
      \makebox{\includegraphics[width = 1.05\columnwidth]{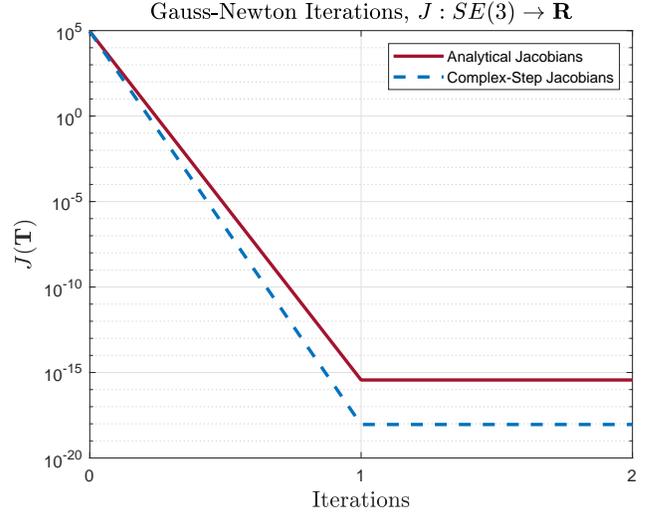}}
      \caption{Convergence history of a Gauss-Newton optimization algorithm on a simple nonlinear least-squares problem. The analytical Jacobians require an approximation to be tractable. The complex-step can calculate Jacobians down to machine precision, hence providing a more accurate first step. \vspace{-0.5cm} }
      \label{fig:exampleGaussNewton}
\end{figure} 
  
   \begin{figure*}[b]
      \centering
      \makebox{\includegraphics[trim = 12cm 9cm 12cm 9.5cm, clip, width = \textwidth]{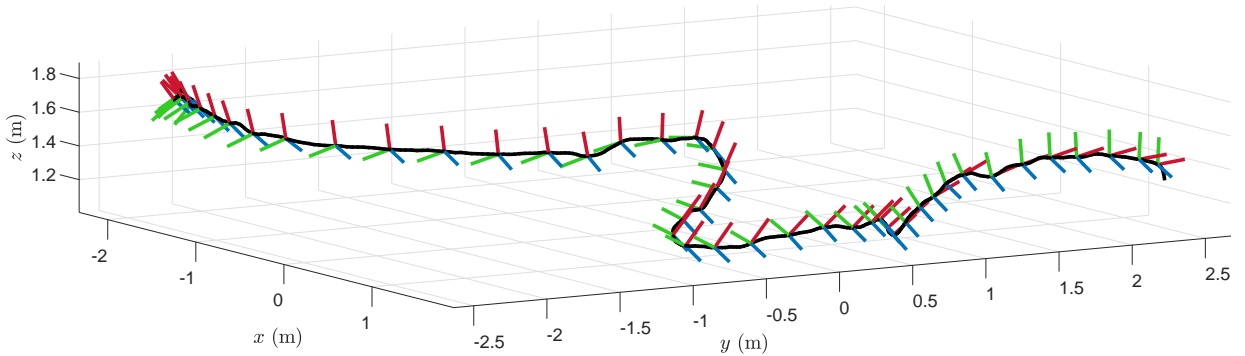}}
      \caption{Trajectory visualization of a batch-estimation solution from the EuRoC Dataset.}
      \label{fig:batchTraj}
   \end{figure*}

\emph{Example 2:} Consider the nonlinear least-squares function
\beq\label{eq:lsLie}
J(\mbf{T}) = \frac{1}{2} \mbf{e}(\mbf{T})^\trans \mbf{W} \mbf{e}(\mbf{T}),
\eeq
where $\mbf{T} = \exp(\mbs{\epsilon}^{L^\wedge}) \mbfbar{T}\in SE(3)$, $\mbf{W} \in \mathbb{R}^{6 \times 6}$ is a symmetric positive definite weight matrix, and the error is given by
\bdis
\mbf{e}(\mbf{T}) = \ln(\mbf{T}^{-1} \mbf{T}^{\mathrm{ref}} )^\vee.
\edis
The matrix $\mbf{T}^{\mathrm{ref}} \in SE(3)$ is some reference point used to construct the error. The Jacobian $\p \mbf{e}(\mbf{T}(\mbs{\epsilon}^L))/ \p \mbs{\epsilon}^L$ can be used to construct Jacobian and Hessian approximations of $J(\mbf{x})$, which are used in the Gauss-Newton algorithm. Like in Example 1, the analytical left Jacobian can be determined by perturbing $\mbf{T}$ on the left,
\begin{align*}
\mbf{e}(\exp(\mbs{\epsilon}^{L^\wedge})\mbfbar{T}) &= \ln(\mbfbar{T}^{-1} \exp(-\mbs{\epsilon}^{L^\wedge})  \mbf{T}^{\mathrm{ref}} )^\vee\\
&= \ln(\exp((-\mathrm{Ad}(\mbfbar{T}^{-1})\mbs{\epsilon}^L)^\wedge) \underbrace{\mbfbar{T}^{-1}\mbf{T}^{\mathrm{ref}}}_{\exp(\mbf{e}(\mbfbar{T})^\wedge)} )^\vee \\
& \approx \mbf{e}(\mbfbar{T}) + \underbrace{(-\mathrm{Ad}(\mbfbar{T}^{-1}))}_{\left.\frac{\p \mbf{e}(\mbf{T}(\mbs{\epsilon}^L))}{ \p \mbs{\epsilon}^L}\right|_{\mbs{\epsilon}^L = \mbf{0}}} \mbs{\epsilon}^L,
\end{align*}
where, in the last line, a first-order approximation to the BCH formula has been used. 

The elements of the Jacobian $\p \mbf{e}(\mbf{T}(\mbs{\epsilon}^L)) / \p \mbs{\epsilon}^L$ were also calculated using \eqref{eq:complex_lie2} with a step size of $h = 10^{-20}$. An optimization was performed with both Jacobian calculation methods, where the Gauss-Newton step $\delta \mbs{\epsilon}_{\ell - 1}$ is is determined from
\bdis
\delta \mbs{\epsilon}_{\ell-1} = \left[ \left(\frac{\p \mbf{e}}{\p \mbs{\epsilon}^L} \right)^\trans \mbf{W} \left(\frac{\p \mbf{e}}{\p \mbs{\epsilon}^L} \right) \right]^{-1} \left[ - \left(\frac{\p \mbf{e}}{\p \mbs{\epsilon}^L} \right)^\trans \mbf{W} \mbf{e}(\mbfbar{T}) \right],
\edis
and the argument of $\mbf{e}(\mbf{T}(\mbs{\epsilon}^L))$ is dropped for conciseness. The point is updated by
\bdis
\mbfbar{T}_{\ell} = \exp(\delta \mbs{\epsilon}_{\ell-1}^\wedge)\mbfbar{T}_{\ell-1}.
\edis

As shown in Fig. \ref{fig:exampleGaussNewton} using both an analytic Jacobian or a complex-step Jacobian results in an optimum being reached in a single step.  Note that calculating Jacobians using the complex-step is shown to have a minor improvement in cost function reduction as compared to the analytical method. The reason is that the analytical method uses a first-order BCH approximation, which is ultimately slightly less accurate than the machine-precision complex-step Jacobian calculations.

\section{Batch Estimation}
The methodology of Example 2 is now applied to a practical state estimation problem. Consider the task of estimating the position and attitude of a rigid body at different points in time $t_0, t_1,\ldots, t_K$ using various measurements. The state of the rigid body at a discrete point in time $t_k$ can be represented by the matrix Lie group element $\mbf{T}_k \in \mc{G}$, where $\mc{G}$ will depend on the estimation task.

\subsection{Maximum A Posteriori Estimation}
The MAP approach \cite[Ch.~8.2.5]{Barfoot2019} to estimate the states in a batch framework results in the minimization of the least-squares cost function shown in \eqref{eq:lsLie}, where the errors to be minimized are
\bdis
\mbf{e}(\mbf{T}_0,\mbf{T}_1,\ldots,\mbf{T}_K) = \bma{c} \mbf{e}_{u,0} \\ \mbf{e}_{u,1} \\ \vdots \\ \mbf{e}_{u,K} \\ \mbf{e}_{y,0} \\ \vdots \\ \mbf{e}_{y,K} \ema.
\edis
The error term $\mbf{e}_{u,0}$ represents an error between the known initial state $\check{\mbf{T}}_0$, with uncertainty, and the estimated initial state $\mbf{T}_0$. This term is computed as
\bdis
\mbf{e}_{u,0} = \ln(\mbf{T}_0^{-1}\check{\mbf{T}}_0)^\vee.
\edis
The process error terms $\mbf{e}_{u,1}, \dots, \mbf{e}_{u,K}$ are a function of a discrete-time process model of the form ${\mbf{T}_k = \mbf{F}(\mbf{T}_{k-1}, \mbf{u}_{k-1}, \mbf{w}_{k-1})}$ where $\mbf{u}_{k-1}$ and $\mbf{w}_{k-1}$ are the input and zero-mean process noise at time $t_{k-1}$, respectively. These error terms are calculated as
\bdis
\mbf{e}_{u,k} = \ln(\mbf{T}_k^{-1}  \mbf{F}(\mbf{T}_{k-1}, \mbf{u}_{k-1},\mbf{0}))^\vee.
\edis
Finally, the terms $\mbf{e}_{y,0}, \dots, \mbf{e}_{y,K}$ correspond to the errors between measurements, and a measurement model of the form $\mbf{y}_k = \mbf{g}(\mbf{T}_k,\mbs{\nu}_k)$, where $\mbs{\nu}_k$ is zero-mean measurement noise. Hence, the measurement errors are
\bdis
\mbf{e}_{y,k} = \mbf{y}_k - \mbf{g}(\mbf{T}_k,\mbf{0}).
\edis
Following the MAP formulation the weight in \eqref{eq:lsLie} is
\bdis
\mbf{W} = \mathrm{diag}(\mbf{P}_0^{-1}, \mbf{Q}_1^{-1}, \ldots, \mbf{Q}_K^{-1}, \mbf{R}_0^{-1}, \ldots, \mbf{R}_K^{-1}),
\edis
where the matrix $\mbf{P}_0$ is a covariance matrix associated with the uncertainty in the initial state, $\check{\mbf{T}}_0$. The matrices $\mbf{Q}_k$ and $\mbf{R}_k$ are covariance matrices associated with the process and measurement noises, respectively.

The goal is to find  $\mbf{T}_0, \ldots, \mbf{T}_K$ that minimize the least-squares cost function given by \eqref{eq:lsLie}. To use a Gauss-Newton algorithm, the right (or left) Jacobian $\p \mbf{e}(\mbf{T}(\mbs{\epsilon}^R))/ \p \mbs{\epsilon}^R$ is needed, where $\mbs{\epsilon}^R = [\mbs{\epsilon}_0^{R^\trans} \ldots \mbs{\epsilon}_K^{R^\trans}]^\trans$ is a matrix that consists of perturbations to the individual estimated states. Since the error $\mbf{e}(\mbf{T}_0, \ldots , \mbf{T}_K)$ is a function of $K$ different Lie group elements, it is worth mentioning a simple technique that allows a user to treat the same function as a function of a single matrix Lie group element, as shown next in Section \ref{sec:recast}. The Jacobians can then be computed using  \eqref{eq:complex_lie} or \eqref{eq:complex_lie2}.

\subsection{Recasting $f(\mbf{X}_0 ,\dots ,\mbf{X}_K)$ as $f(\mbf{X})$} \label{sec:recast}
Consider a function ${f(\mbf{X}_0,\ldots, \mbf{X}_K) \in \mathbb{R}}$ where ${\mbf{X}_0,\ldots, \mbf{X}_K \in \mc{G}}$. Let ${\mbf{X}_i =  \mbfbar{X}_i\exp(\mbs{\epsilon}_i^{R^\wedge})}$. Define ${ \mbf{X} \isdef \mathrm{diag}(\mbf{X}_0 , \ldots, \mbf{X}_K)}$, thus leading to
\begin{align}
\mbf{X} &= \bma{ccc} \mbfbar{X}_0 && \\& \ddots & \\ && \mbfbar{X}_K \ema \bma{ccc} \exp(\mbs{\epsilon}_0^{R^\wedge}) && \\& \ddots & \\ && \exp(\mbs{\epsilon}_K^{R^\wedge})\ema \nonumber \\
&= \bma{ccc} \mbfbar{X}_0 && \\& \ddots & \\ && \mbfbar{X}_K \ema \exp\bma{ccc} \mbs{\epsilon}_0^{R^\wedge} && \\& \ddots & \\ && \mbs{\epsilon}_K^{R^\wedge}\ema. \label{eq:recast1}
\end{align}
By defining $\mbs{\epsilon}^{R} \isdef \left[ \mbs{\epsilon}_0^{R^\trans} \ldots \mbs{\epsilon}_K^{R^\trans} \right]^\trans$, $ \mbfbar{X} \isdef \mathrm{diag}(\mbfbar{X}_0 , \ldots, \mbfbar{X}_K)$ along with a new operator $(\cdot)^\triangle$ such that ${\mbs{\epsilon}^\triangle \isdef \mathrm{diag}( \mbs{\epsilon}_0^\wedge,\ldots,\mbs{\epsilon}_K^\wedge)}$, equation \eqref{eq:recast1} becomes
\bdis
\mbf{X} =  \mbfbar{X}\exp(\mbs{\epsilon}^{R^\triangle}).
\edis
Therefore, a collection of matrix Lie group elements can be packaged into a single element of a new group. This can be done similarly with left perturbations. 

   \begin{figure}[t]
      \centering
      \makebox{\includegraphics[width = 1.05\columnwidth]{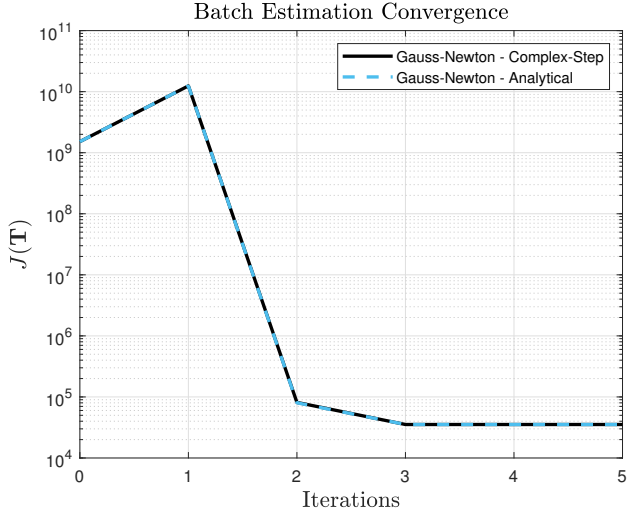}}
      \caption{Convergence history of a Gauss-Newton algorithm on EuRoC Dataset. Virtually identical performance is achieved to the analytical solution, and using a central-difference method. However, the complex-step requires only 1 function evaluation, and no step-size tuning was needed.\vspace{-0.5cm} }
      \label{fig:batchETHconvergence}
   \end{figure}  

\begin{figure}[t]
      \centering
      \makebox{\includegraphics[width = 1.05\columnwidth]{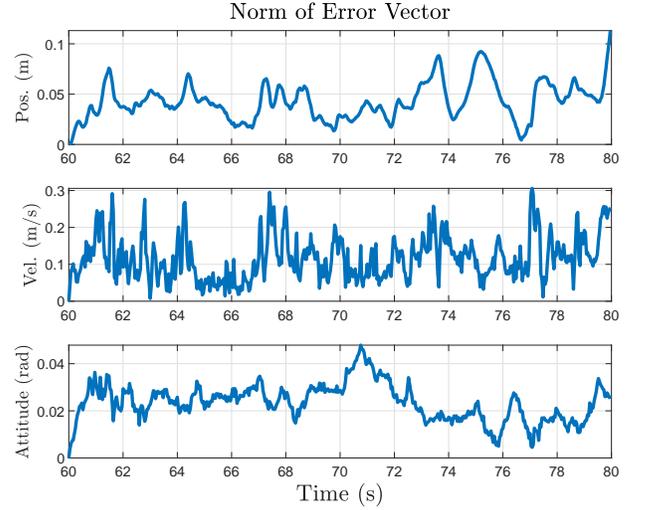}}
      \caption{Magnitude of errors in position, velocity, and attitude resulting from the optimal batch-estimation solution using the complex-step.\vspace{-0.5cm} }
      \label{fig:normError}
\end{figure} 
 
\subsection{The EuRoC Dataset}
 
The EuRoC micro aerial vehicle dataset collected by the Autonomous Systems Laboratory at ETH Z\"{u}rich, Switzerland \cite{Burri2016} includes accelerometer and gyroscope measurements, as well as ground truth position data. To simulate position measurements akin to GPS or UWB measurements, normally distributed random noise is added to the provided ground position data. The state of the rigid body can be represented by the matrix Lie group element ${\mbf{T}_k \in SE_2(3)}$, and as such the velocity is also estimated. The accelerometer measurements $\mbf{u}^\mathrm{acc}_k$ and gyroscope measurements $\mbf{u}^\mathrm{gyro}_k$ are treated as process-model inputs ${ \mbf{u}_k = [\mbf{u}^{\mathrm{acc}^\trans}_{k} \;\mbf{u}^{\mathrm{gyro}^\trans}_{k}]^\trans}$, while the position measurements $\mbf{y}^{\mathrm{pos}}_k$ are treated as measurement-model outputs.

For this problem, the analytical expression for the Jacobian $\p \mbf{e}(\mbf{T}) /\p \mbs{\epsilon}^R$ can be obtained, and the details of the derivation can be found in \cite[Ch.~5]{Arsenault2019}. Henceforth, the arguments of functions of multiple matrix Lie group elements will be consolidated under $\mbf{T}$, as described in Section \ref{sec:recast}. The right Jacobian is
\bdis
\frac{\p \mbf{e} (\mbf{T})}{\p \mbs{\epsilon}^R} \approx \bma{ccc} -\mbf{1} & & \\ \mbf{F}_0 & \ddots & \\ & \ddots & \\ && -\mbf{1} \\ && \mbf{F}_K \\ \hline \mbf{H}_0 && \\  &\ddots &\\ && \mbf{H}_K \ema,
\edis
where
\begin{align*}
\mbf{F}_k &= \mathrm{Ad}(\mbf{T}_k^{-1}\mbf{F}^\mathrm{op}_{k-1})\mbf{B},\\ 
\mbf{F}^\mathrm{op}_{k-1} &= \bma{ccc} \mbf{C}_{k-1} & \mbf{v}_{k-1}  + T\mbf{g} & \mbf{r}_{k-1}  + T\mbf{v}_{k-1}  \\ & 1 & \\ &&1 \ema ,\\
\mbf{B} &= \bma{ccc} \mbf{1} && \\ &\mbf{1}& \\ & T\mbf{1}& \mbf{1} \ema,\\
\mbf{H}_k &= \bma{ccc}\mbf{1} & \mbf{0} & \mbf{0}\ema \mbf{T}_k \mbf{p}^\odot ,\\
\end{align*}
where $T = t_k - t_{k-1}$, $\mbf{p} = [\mbf{0} \; 1]^\trans$, and $\mbf{g}$ is the gravity vector resolved in the datum frame. These expressions require  first-order approximations to the BCH formula, similar to Example 2. This is common procedure, as the approximation becomes more accurate as errors become small \cite{Arsenault2019, Barfoot2019}.

A Gauss-Newton optimization is performed on the {\texttt{MH\_03\_medium}} dataset. For simplicity, the accelerometer and gyroscope measurements are downsampled from the original {200 Hz} in order to reduce the amount of variables in the optimization procedure. An alternative to downsampling is to perform IMU preintegration as described in \cite{Forster2017}, but this is beyond the scope of this paper. The specifications of the batch-estimation problem are shown in Table \ref{table:batchSpecs}. The process covariance matrix was set to,
\beq \label{eq:proc_cov}
\mbf{Q}_k = \mathrm{diag}( 1.6 \cdot 10^{-7} \cdot\mbf{1}\; ,\;  2 \cdot 10^{-6} \cdot\mbf{1}\; ,\; 10^{-10} \mbf{1}).
\eeq
  
 \vspace{-0.3cm}
\begin{table}[h]
\caption{EuRoC Estimation Scenario Specifications}
\vspace{-0.5cm}
\begin{center}
\begin{tabular}{c|c|c}
\hline
\textbf{Specification} & \textbf{Value} & \textbf{Units}\\
\hline \hline
Accelerometer meas. freq. & 25 & Hz \\
Gyroscope meas. freq. & 25 & Hz \\
Position meas. freq. & 10 & Hz \\
Data time span & 60 - 80 & s \\
Number of states estimated & 500 & - \\
Std. deviation of position meas. & 0.1 & m \\
Initial state guess covariance $\mbf{P}_0$ & $ 10^{-10}\cdot \mbf{1} $ & [rad$^2$, (m/s)$^2$, m$^2$]\\
Process covariance $\mbf{Q}_k$ & See eqn. \eqref{eq:proc_cov} & [rad$^2$, (m/s)$^2$, m$^2$]\\
Measurement covariance $\mbf{R}_k$ & $0.1^2\cdot\mbf{1} $ & m$^2$\\
Complex-step der. step size $h$ & $10^{-20}$ & -\\
\hline
\end{tabular}
\label{table:batchSpecs}
\end{center}
\end{table}
\vspace{-0.2cm}

\noindent
The initial state, $\check{\mbf{T}}_0$, is set to the ground truth, and hence the diagonal of $\mbf{P}_0$ is given arbitrarily small numbers. The matrix $\mbf{Q}_k$ was further tuned to yield better performance, after obtaining the nominal noise values provided in the EuRoC dataset. Using the initial state, the process model is directly integrated using the accelerometer and gyroscope measurements, which then provides an initial guess for the poses at all the discrete time points. This dead reckoning solution is then used to initialize the Gauss-Newton algorithm.

Figure~\ref{fig:batchTraj} shows a visualization of the trajectory once the optimization procedure has converged. Figure~\ref{fig:batchETHconvergence} shows the value of the cost function $J(\mbf{T})$ across the iterations of the Gauss-Newton algorithm. Since the initial guess for the states is obtained by dead reckoning, this sets all the process errors $\mbf{e}_{u,1},\ldots,\mbf{e}_{u,K}$ to zero. The first iteration attempts to decrease the measurement errors, resulting in an increase in process errors, and hence an increase in the overall cost function.

  In this example, BCH approximations in the analytical Jacobians did not create any difference in the convergence history since the errors are initialized to be small in the the dead reckoning step. A central-difference scheme was also used to calculate Jacobians, and after multiple trial-and-error attempts with different step sizes, an identical convergence history to what is shown in Fig. \ref{fig:batchETHconvergence} was obtained. However, the central-difference method requires twice as many function evaluations as the complex-step method, and therefore required approximately twice the total computing time. Finally, Fig.~\ref{fig:normError} shows the 2-norm of the difference between the batch-estimation solution and the ground truth. The errors are small, indicating the MAP framework has converged close to the ground truth.

\subsection{The `Lost in the Woods' Dataset}
\begin{figure}[t]
      \centering
      \makebox{\includegraphics[width = \columnwidth]{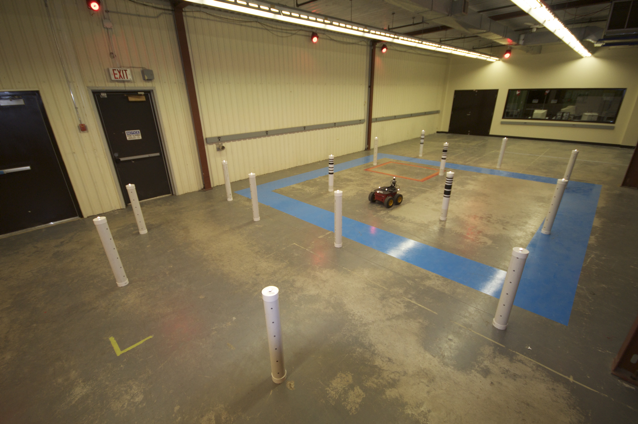}}
      \caption{Experimental setup of the `Lost in the Woods' dataset, courtesy of \cite{Barfoot2009}. Truth measurements are obtained from a motion capture system. \vspace{-0.5cm} }
      \label{fig:woods}
\end{figure}

The `Lost in the Woods' dataset consists of a mobile wheeled robot navigating through a ``forest'' of tubes \cite{Barfoot2009}, as seen in Figure \ref{fig:woods}. The robot is equipped with wheel odometry providing forward velocity measurements, denoted $u^{\mathrm{vel}}_k$, and angular velocity measurements, denoted $u^{\mathrm{ang}}_k$. Furthermore, the robot has a laser range finder that provides range and bearing measurements to pre-identified landmarks (the tubes shown in Figure \ref{fig:woods}), denoted $r^{\ell}_k, \phi^\ell_k$ for landmark $\ell$ at $t_k$, respectively. The positions of the landmarks in a datum reference frame are known in advance, and are denoted $\mbf{r}^\ell$. The state of the robot can be represented by $\mbf{T}_k \in SE(2)$.

  \vspace{-0.2cm}
\begin{table}[h]
\caption{`Lost in the Woods' Estimation Scenario Specifications}
\vspace{-0.2cm}
\begin{center}
\begin{tabular}{c|c|c}
\hline
\textbf{Specification} & \textbf{Value} & \textbf{Units}\\
\hline \hline
Wheel odometry freq. & 5 & Hz \\
Laser range finder freq. & 5 & Hz \\
Data time span & 500 - 620 & s \\
Number of states estimated & 600 & - \\
Initial state guess covariance $\mbf{P}_0$ & $\mbf{1} $ & [rad$^2$, $m^2$, m$^2$]\\
Complex-step der. step size $h$ & $10^{-20}$ & -\\
\hline
\end{tabular}
\label{table:woodsSpecs}
\end{center}
\end{table}
\noindent
\vspace{-0.3cm}

The process model consists of the nonholonomic vehicle kinematics. Written in the form ${\mbf{T}_k = \mbf{F}(\mbf{T}_{k-1}, \mbf{u}_{k-1}, \mbf{w}_{k-1})}$,
\bdis
\mbf{T}_k = \mbf{T}_{k-1} \mbs{\Psi}_{k-1},
\edis 
where
\bdis
\mbs{\Psi}_{k-1} = \bma{cc} \exp(T(u^{\mathrm{ang}}_{k-1} +w^{\mathrm{ang}}_{k-1})^\wedge) & T(u^{\mathrm{vel}}_{k-1}+w^{\mathrm{vel}}_{k-1})\mbf{1}_{1} \\ \mbf{0} & 1 \ema ,
\edis
$T = t_k - t_{k-1}$, and $w^{\mathrm{vel}}_{k-1},w^{\mathrm{ang}}_{k-1}$ are zero-mean normally distributed noises associated with the velocity and angular velocity measurements, respectively. The measurement model consists of the range and bearing measurements for each landmark. Written as $\mbf{y} = \mbf{g}(\mbf{T}_k,\mbs{\nu}_k)$, the measurement model is
\bdis
\bma{c} r^\ell_k \\ \phi^\ell_k \ema = \bma{c} \sqrt{(\mbf{r}^\ell - \mbf{D}\mbf{T}_k\mbf{p})^\trans(\mbf{r}^\ell - \mbf{D}\mbf{T}_k\mbf{p})} \\ \hspace{-8pt}
 \left( \mathrm{atan2} \left( \mbf{1}_2^\trans(\mbf{r}^\ell - \mbf{D}\mbf{T}_k\mbf{p}), \mbf{1}_1^\trans (\mbf{r}^\ell - \mbf{D}\mbf{T}_k\mbf{p}) \right) \right. \hspace{-10pt} \\ \hspace{3.6cm} \left. - \mbf{1}_1^\trans \ln (\mbf{T}_k)^{\vee} \right) \hspace{-6pt}\ema + \mbs{\nu}_k,
\edis
where $\mbs{\nu}_k$ is zero-mean normally distributed measurement noise, $\mbf{D} = [\mbf{1} \; \mbf{0}]$, $\mbf{p} = [d \; 0 \; 1]^\trans$, and $d$ is the distance between the laser range finder and the reference point on the robot.

Computing the Jacobians associated with the measurement model by hand is, although not impossible, laborious due to the $\mathrm{atan2(\cdot,\cdot)}$ term. Hence, the complex-step derivative is used to directly evaluate the right Jacobian $\p \mbf{e}(\mbf{T}) /\p \mbs{\epsilon}^R$ for use in the Gauss-Newton optimization. 

Dead reckoning was performed using wheel odometry in order to generate an initial guess for the Gauss-Newton optimization. All measurements were downsampled from the original 10 Hz to 5 Hz in order to limit the number of variables in the optimization procedure. The $\mbf{Q}_k$ and $\mbf{R}_k$ matrices were directly formed from the discrete-time covariances provided in the dataset \cite{Barfoot2009}. The initial state $\mbf{T}_0$ was set to be a random perturbation from the ground truth.

\begin{figure}[t]
      \centering
      \makebox{\includegraphics[width = 1.05\columnwidth]{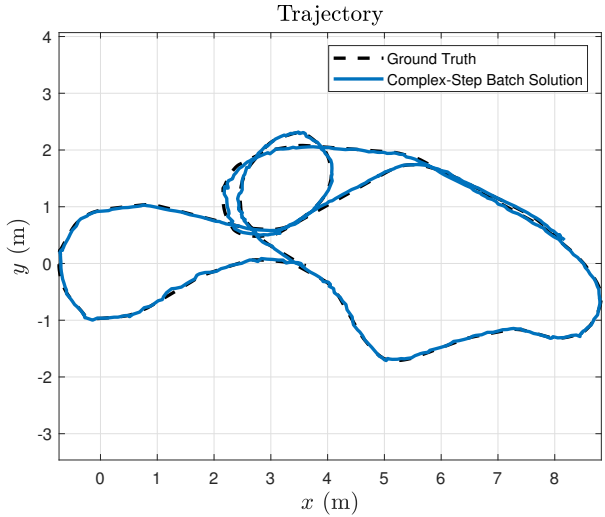}}
      \caption{2D trajectory trace for the `Lost in the Woods' dataset. The solution using the complex-step derivative shows excellent agreement with the ground truth data. \vspace{-0.5cm} }
      \label{fig:barfootTraj}
\end{figure}

\begin{figure}[t]
      \centering
      \makebox{\includegraphics[width = 1.05\columnwidth]{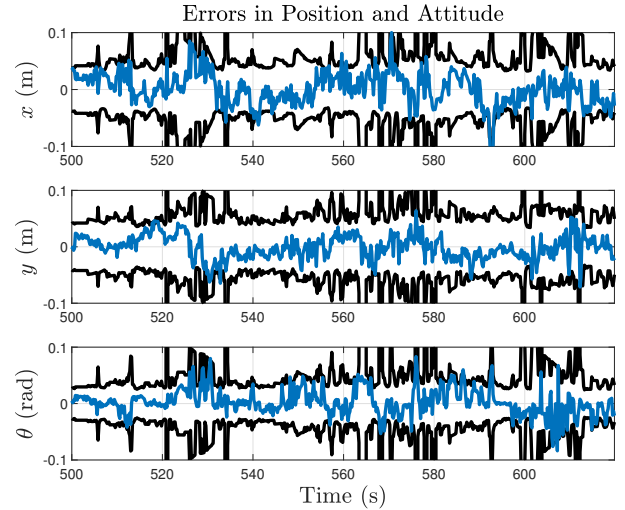}}
      \caption{Error in position $x,y$ and attitude $\theta$ between estimated solution and ground truth (blue), along with $\pm 3$ standard deviation bounds (black). There is less than 10 cm of position error, and less than 0.1 rad of attitude error. \vspace{-0.5cm} }
      \label{fig:barfootError}
\end{figure}

 The algorithm converged in 6 iterations, and produced a trajectory visualizable in Figure \ref{fig:barfootTraj}. The errors are shown in Figure \ref{fig:barfootError}, which show good performance when compared to the ground truth position and attitude data. This can also be achieved with central-difference, but again, the computation time is significantly longer, and the step sized must be tuned.

\section{Conclusion}
This paper has shown that the complex-step derivative can successfully be used to obtain Jacobians of functions that have matrix Lie group elements as arguments. Machine-precision can be achieved with a single complex function evaluation. To use the complex-step, functions must be programmed to accept complex numbers, which is occasionally time consuming. In \textsc{Matlab}, it is critical to use the (\texttt{.'}) transpose operator as opposed to the (\texttt{'}) conjugate transpose, and also to redefine the \texttt{abs()}, \texttt{max()}, and \texttt{min()} functions. A guide to proper implementation in various other programming languages can be found in \cite{MDOUMich2017}.

There is a multitude of other potential applications for this tool, such as numerical linearization of high-fidelity dynamics models, real-time state estimation and Kalman filtering \cite{Vittaldev2012}, and the training of matrix Lie group-based neural networks \cite{Peretroukhin2018}. For second derivatives, the complex-step is unfortunately unable to realize machine-precision accuracy. However, methods are available to improve the accuracy \cite{Lai2008}, which are likely extendible to matrix Lie groups. Furthermore, if an analytical Jacobian is known, the Hessian can be determined with machine precision using the complex-step \cite{Martins2003}.

\appendices
\section*{Appendix}
\subsection{The Special Euclidean Group $SE(2)$}
The group $SE(2)$ is defined as \cite{Chirikjian2009},
\bdis
SE(2) = \left\lbrace \mbf{T} = \bma{cc} \mbf{C} & \mbf{r} \\ \mbf{0} & 1 \ema \in \mathbb{R}^{3 \times 3} \; \bigg \vert \; \mbf{C} \in SO(2) \; , \mbf{r} \in \mathbb{R}^2 \right\rbrace,
\edis
where $SO(n)$ refers to the Special Orthogonal Group consisting of orthonormal matrices with unit determinant. The matrix Lie algebra associated with $SE(2)$ is
 \bdis
 \mathfrak{se}(2) = \{ \mbs{\Xi} = \mbs{\xi}^\wedge \in \mathbb{R}^{3 \times 3} \; | \; \mbs{\xi} \in \mathbb{R}^3 \},
 \edis
 where
  \bdis
\mbs{\xi}^{\wedge} = \bma{c} {\xi}^{\phi} \\ {\xi}^{r}_1 \\ {\xi}^{r}_2 \ema^\wedge = \bma{ccc} 0 & -{\xi}^{\phi} & {\xi}^{r}_1\\ {\xi}^{\phi} & 0 & {\xi}^{r}_2 \\ 0& 0 & 0 \ema.
\edis
The closed-form expression for the exponential map $ {\exp : \mathfrak{se}(2) \to SE(2)}$ is
\bdis
\exp(\mbs{\xi}^\wedge) = \bma{cc}\mbf{C}& \mbf{J}_\ell \mbs{\xi}^{r} \\ \mbf{0} & 1 \ema,
\edis
where $\mbs{\xi}^r = [\xi^r_1 \; \xi^r_2]^\trans$ and 
\bdis
\mbf{J}_\ell = \frac{1}{\xi^\phi}\bma{cc} \sin(\xi^\phi)&-(1 - \cos(\xi^\phi)) \\ (1 - \cos(\xi^\phi)) & \sin(\xi^\phi) \ema.
\edis

\subsection{The Special Euclidean Group $SE(3)$}
 The matrix Lie group $SE(3)$ is defined as \cite[Ch.~7.1.2]{Barfoot2019}
\bdis
SE(3) = \left\lbrace \mbf{T} = \bma{cc} \mbf{C} & \mbf{r} \\ \mbf{0} & 1 \ema \in \mathbb{R}^{4 \times 4} \; \bigg \vert \; \mbf{C} \in SO(3) \; , \mbf{r} \in \mathbb{R}^3 \right\rbrace.
\edis
 The matrix Lie algebra associated with $SE(3)$ is
 \bdis
 \mathfrak{se}(3) = \{ \mbs{\Xi} = \mbs{\xi}^\wedge \in \mathbb{R}^{4 \times 4} \; | \; \mbs{\xi} \in \mathbb{R}^6 \},
 \edis
 where
 \bdis
 \mbs{\xi}^\wedge = \bma{c} \mbs{\xi}^\phi \\ \mbs{\xi}^r \ema^\wedge = \bma{cc} \mbs{\xi}^{\phi^\times} &   \mbs{\xi}^r \\ \mbf{0} & 0 \ema, \qquad \mbs{\xi}^\phi , \mbs{\xi}^r \in \mathbb{R}^3,
 \edis
 and
 \bdis
\mbs{\xi}^{\phi^\times} = \bma{c} {\xi}^{\phi}_1 \\ {\xi}^{\phi}_2 \\ {\xi}^{\phi}_3 \ema^\times = \bma{ccc} 0 & -{\xi}^{\phi}_3 & {\xi}^{\phi}_2 \\ {\xi}^{\phi}_3 & 0 & - {\xi}^{\phi}_1 \\ - {\xi}^{\phi}_2& {\xi}^{\phi}_1& 0 \ema.
\edis
The closed-form expression for the exponential map  ${\exp :\mathfrak{se}(3) \to SE(3)}$ is
\bdis
\exp(\mbs{\xi}^\wedge) = \bma{cc} \exp(\mbs{\xi}^{\phi^\times}) & \mbf{J}_\ell \mbs{\xi}^{r} \\ \mbf{0} & 1 \ema,
\edis
where
\bdis
\mbf{J}_\ell = \frac{\sin(\phi)}{\phi} \mbf{1} + \left(1 - \frac{\sin(\phi)}{\phi} \right) \mbf{a}\mbf{a}^\trans + \frac{1 - \cos(\phi)}{\phi} \mbf{a}^\times, \edis \bdis
\exp (\mbs{\xi}^{\phi^\times}) = \cos (\phi)\mbf{1} + ( 1 - \cos (\phi)) \mbf{aa}^\trans + \sin (\phi) \mbf{a}^\times,
\edis
and  $\phi = \norm{\mbs{\xi}^\phi}$ and $\mbf{a} = \mbs{\xi}^\phi/\phi$. The matrix $\mbf{J}_\ell$ is known as the {left Jacobian} of the group $SO(3)$. It is also useful to define the operator \cite[Ch.~7.1.8]{Barfoot2019}
\bdis
\mbf{p}^\odot = \bma{c} \mbs{\varepsilon} \\ \eta \ema^\odot = \bma{cc} - \mbs{\varepsilon}^\times & \eta \mbf{1} \\ \mbf{0} & \mbf{0} \ema, \qquad \mbs{\varepsilon} \in \mathbb{R}^3, \eta \in \mathbb{R},
\edis
such that $\mbf{x}^\wedge \mbf{p} = \mbf{p}^\odot \mbf{x}$ holds. 

\subsection{The Group of Double Direct Isometries $SE_2(3)$}
The matrix Lie group $SE_2(3)$ is defined as
\bdis
SE_2(3) = \left\lbrace \mbf{T} = \bma{ccc} \mbf{C} & \mbf{v}& \mbf{r} \\ \mbf{0} & 1 & 0 \\ \mbf{0} & 0& 1 \ema  \bigg \vert \; \mbf{C} \in SO(3) ,\;  \mbf{v},\mbf{r} \in \mathbb{R}^3 \right\rbrace.
\edis
 The matrix Lie algebra associated with $SE_2(3)$ is
 \bdis
 \mathfrak{se}_2(3) = \{ \mbs{\Xi} = \mbs{\xi}^\wedge \in \mathbb{R}^{5 \times 5} \; | \; \mbs{\xi} \in \mathbb{R}^9 \},
 \edis
 where
 \bdis
 \mbs{\xi}^\wedge = \bma{c} \mbs{\xi}^\phi \\ \mbs{\xi}^v\\ \mbs{\xi}^r \ema^\wedge = \bma{ccc} \mbs{\xi}^{\phi^\times} &   \mbs{\xi}^v &   \mbs{\xi}^r \\ \mbf{0} & 0& 0\\ \mbf{0} & 0& 0 \ema, \quad \mbs{\xi}^\phi , \mbs{\xi}^v ,\mbs{\xi}^r \in \mathbb{R}^3.
 \edis
The closed-form expression for the exponential map $ {\exp : \mathfrak{se}_2(3) \to SE_2(3)}$ is
\bdis
\exp(\mbs{\xi}^\wedge) = \bma{ccc} \exp(\mbs{\xi}^{\phi^\times}) & \mbf{J}_\ell \mbs{\xi}^{v}& \mbf{J}_\ell \mbs{\xi}^{r} \\ \mbf{0} & 1 & 0\\ \mbf{0} & 0 & 1\ema.
\edis
It is also useful to define the operator
\bdis
\mbf{p}^\odot = \bma{c} \mbs{\varepsilon} \\ \eta_1 \\ \eta_2 \ema^\odot = \bma{ccc} - \mbs{\varepsilon}^\times & \eta_1 \mbf{1} & \eta_2 \mbf{1} \\ \mbf{0} & \mbf{0} & \mbf{0}  \ema ,
\edis
where $\mbs{\varepsilon} \in \mathbb{R}^3$ and  $\eta_1,\eta_2 \in \mathbb{R}$, such that $\mbf{x}^\wedge \mbf{p} = \mbf{p}^\odot \mbf{x}$ holds.

\section*{Acknowledgment}
The authors would like to thank Jonathan Arsenault and Thomas Hitchcox for their many helpful discussions.

\ifCLASSOPTIONcaptionsoff
  \newpage
\fi



%

%
%

{\AtNextBibliography{\small}
\printbibliography}

@inproceedings{Barfoot2009,
address = {Toronto, ON},
author = {Barfoot, Timothy David},
booktitle = {State Estimation for Aerospace Vehicles},
publisher = {Uni. of Toronto},
title = {{AER 1513 Course Assignments}},
year = {2009}
}

@book{Chirikjian2009,
author = {Chirikjian, Gregory S.},
publisher = {Birkh{\"{a}}user Boston},
title = {{Stochastic Models, Information Theory, and Lie Groups - Volume 2}},
year = {2009}
}

@article{Martins2003,
author = {Martins, Joaquim R R A and Sturdza, Peter and Alonso, Juan J},
journal = {ACM Trans. on Mathematical Software},
number = {3},
pages = {245--262},
title = {{The Complex-Step Derivative Approximation}},
volume = {29},
year = {2003}
}

@misc{Ceres2019,
author = {Agarwal, Sameer and Mierle, Keir},
title = {{Ceres Solver — A Large Scale Non-linear Optimization Library}},
url = {http://ceres-solver.org/},
urldate = {2019-11-19},
year = {2019}
}

@article{Boumal2014,
author = {Boumal, Nicolas and Mishra, Bamdev and Absil, P.-A and Sepulchre, Rodolphe},
journal = {Machine Learning Research},
pages = {1455--1459},
title = {{Manopt, a Matlab Toolbox for Optimization on Manifolds}},
volume = {15},
year = {2014}
}

@article{Peretroukhin2018,
author = {Peretroukhin, Valentin and Kelly, Jonathan},
journal = {IEEE Robotics and Automation Letters},
number = {3},
pages = {2424--2431},
title = {{DPC-Net: Deep Pose Correction for Visual Localization}},
volume = {3},
year = {2018}
}

@incollection{Sommer2013,
author = {Sommer, Hannes and Pradalier, C{\'{e}}dric and Furgale, Paul},
booktitle = {Robotics Research: The 16th International Symposium},
pages = {505--520},
title = {{Automatic Differentiation on Differentiable Manifolds as a Tool for Robotics}},
year = {2013}
}

@article{Burri2016,
author = {Burri, Michael and Nikolic, Janosch and Gohl, Pascal and Schneider, Thomas and Rehder, Joern and Omari, Sammy and Achtelik, Markus W and Siegwart, Roland},
journal = {The International Journal of Robotics Research},
number = {10},
pages = {1157--1163},
title = {{The EuRoC Micro Aerial Vehicle Datasets}},
volume = {35},
year = {2016}
}

@article{Vittaldev2012,
author = {Vittaldev, Vivek and Russell, Ryan P and Arora, Nitin and Gaylor, David},
journal = {Advances in the Astronautical Sciences},
number = {1},
pages = {1--16},
title = {{Second-Order Kalman Filters Using Multi-Complex Step Derivatives}},
volume = {143},
year = {2012}
}

@article{Lai2008,
author = {Lai, K.-L and Crassidis, J L},
journal = {J. Comp. Applied Math},
number = {1},
pages = {276--293},
title = {{Extensions of the First and Second Complex-Step Derivative Approximations}},
volume = {219},
year = {2008}
}

@misc{GTSAM2019a,
author = {GTSAM},
pages = {1--29},
title = {{Math of GTSAM}},
url = {https://github.com/borglab/gtsam},
urldate = {2019-09-03},
year = {2019}
}

@book{Absil2008,
author = {Absil, P.A. and Mahony, R. and Sepulchre, R.},
publisher = {Princeton Uni. Press},
title = {{Optimization Algorithms on Matrix Manifolds}},
year = {2008}
}

@article{Townsend2016,
author = {Townsend, James and Weichwald, Sebastian},
journal = {Journal of Machine Learning Research},
number = {1},
pages = {1--5},
title = {{Pymanopt: A Python Toolbox for Optimization on Manifolds using Automatic Differentiation}},
volume = {17},
year = {2016}
}

@phdthesis{Arsenault2019,
author = {Arsenault, Jonathan},
number = {August},
school = {McGill University},
title = {{Practical Considerations and Extensions of the Invariant Extended Kalman Filtering Framework}},
type = {M.A.Sc. Thesis},
year = {2019}
}

@misc{Sophus2019,
author = {Strasdat, Hauke},
title = {{Sophus - Lie groups for 2D/3D Geometry}},
url = {https://strasdat.github.io/Sophus/},
urldate = {2019-11-19},
year = {2019}
}

@book{Hall2015,
address = {New York, NY},
author = {Hall, Brian},
edition = {second},
publisher = {Springer},
title = {{Lie Groups, Lie Algebras and Representations}},
year = {2015}
}

@article{Squire1998,
author = {Squire, William and Trapp, George},
journal = {SIAM Review},
number = {1},
pages = {110--112},
title = {{Using Complex Variables to Estimate Derivatives of Real Functions}},
volume = {40},
year = {1998}
}

@article{Forster2017,
author = {Forster, Christian and Carlone, Luca and Dellaert, Frank and Scaramuzza, Davide},
journal = {IEEE Trans. Robotics},
number = {1},
pages = {1--21},
title = {{On-Manifold Preintegration for Real-Time Visual-Inertial Odometry}},
volume = {33},
year = {2017}
}

@book{Barfoot2019,
address = {Toronto, ON},
author = {Barfoot, Tim},
publisher = {Cambridge University Press},
title = {{State Estimation for Robotics}},
year = {2019}
}

@misc{MDOUMich2017,
author = {{MDO Laboratory}},
booktitle = {University of Michigan},
title = {{A Guide to the Complex-Step Derivative Approximation}},
url = {http://mdolab.engin.umich.edu},
urldate = {2019-10-31},
year = {2017}
}

@inproceedings{Robenack2011,
author = {R{\"{o}}benack, Klaus and Winkler, Jan and Wang, Siqian},
booktitle = {Workshop on Equation-Based Object-Oriented Modeling Languages and Tools},
organization = {ETH Z{\"{u}}rich},
pages = {57--66},
title = {{LIEDRIVERS - A Toolbox for the Efficient Computation of Lie Derivatives Based on the Object-Oriented Algorithmic Differentiation Package ADOL-C}},
year = {2011}
}

%








\end{document}